\pdfoutput=1

\documentclass[runningheads]{llncs}
\usepackage{graphicx}
\usepackage{amsmath,amssymb} 
\usepackage{color}

\usepackage{booktabs}
\usepackage{cite}
\usepackage{mmstyles}

\begin{document}
\title{Move Forward and Tell: A Progressive Generator of Video Descriptions} 

\titlerunning{Move Forward and Tell: A Progressive Generator of Video Descriptions}
\author{Yilei Xiong \and
Bo Dai \and
Dahua Lin}
\authorrunning{Yilei Xiong, Bo Dai, Dahua Lin}

\institute{CUHK-SenseTime Joint Lab,
	The Chinese University of Hong Kong
	\email{ \{xy014,db014,dhlin\}@ie.cuhk.edu.hk}
}
\maketitle              
%


\begin{abstract}
We present an efficient framework that can generate a coherent paragraph to
describe a given video.
Previous works on video captioning usually focus on video clips.
They typically treat an entire video as a whole and generate the caption
conditioned on a single embedding.
On the contrary, we consider videos with rich temporal structures and
aim to generate paragraph descriptions that can preserve the story flow
while being coherent and concise.
Towards this goal, we propose a new approach, which produces a descriptive
paragraph by assembling temporally localized descriptions.
Given a video, it selects a sequence of distinctive clips
and generates sentences thereon in a coherent manner.
Particularly, the selection of clips and the production of sentences
are done jointly and progressively driven by a recurrent network
-- what to describe next depends on what have been said before.
Here, the recurrent network is learned via self-critical sequence training
with both sentence-level and paragraph-level rewards.
On the ActivityNet Captions dataset, our method demonstrated the capability
of generating high-quality paragraph descriptions for videos.
Compared to those by other methods, the descriptions produced by our method
are often more relevant, more coherent, and more concise.

\end{abstract}


\section{Introduction}
\label{sec:intro}

\begin{figure*}
	\centering
	\includegraphics[width=\textwidth]{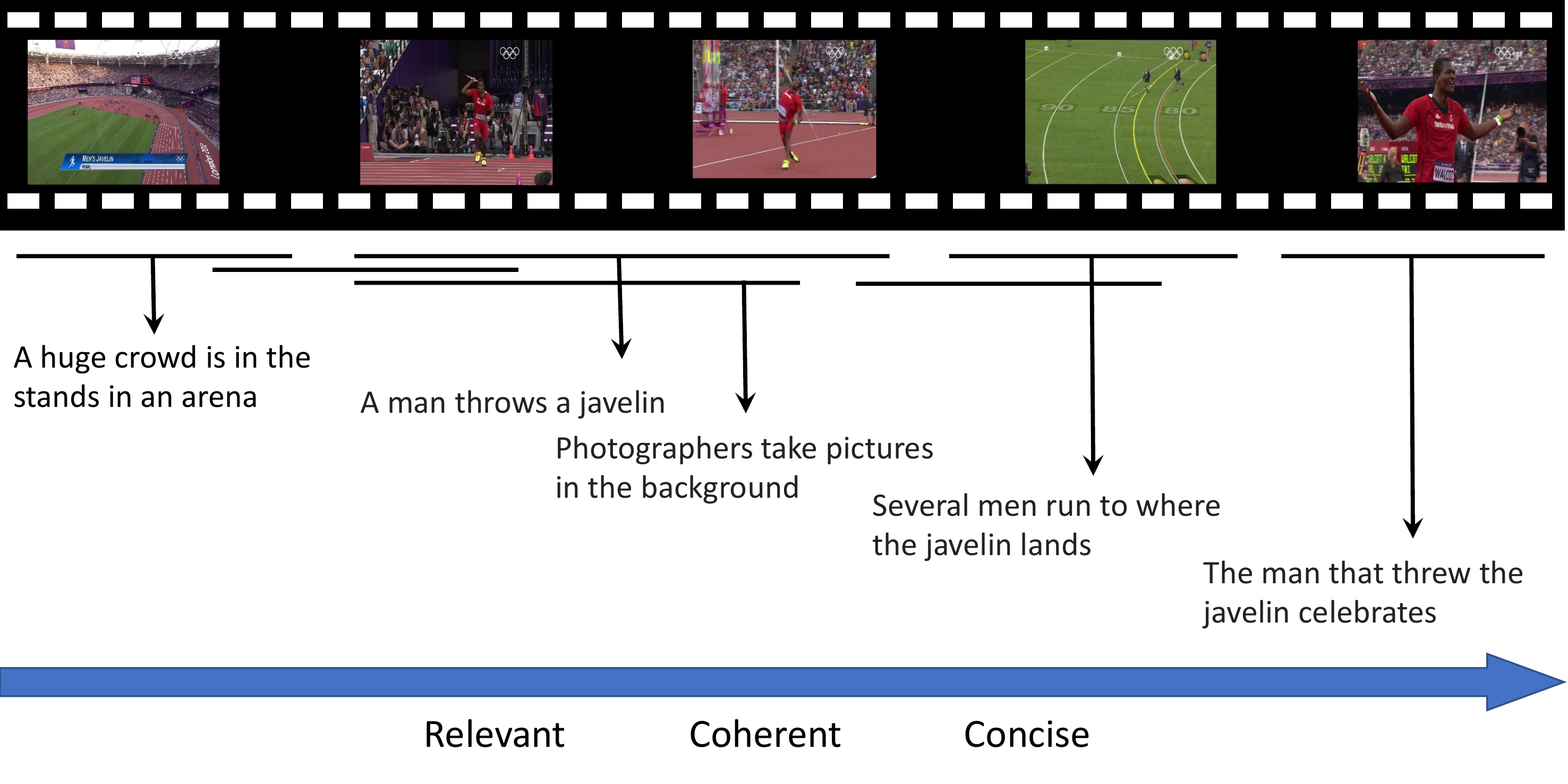}
	\caption{As shown in this figure, our framework localizes important events in a video,
	and picks a sequence of coherent and independent events,
	on which generates a coherent and concise descriptive paragraph for the video.}
	\label{fig:teaser}
\end{figure*}


Textual descriptions are an important way to characterize images
and videos. Compared to class labels or semantic tags, descriptions
are usually more informative and distinctive.
In recent years, image captioning, a task to generate short descriptions
for given images, becomes an active research topic~\cite{baraldi2016hierarchical, yu2017end, hori2017attention, wang2016multimodal} and has seen
remarkable progress thanks to the wide adoption of recurrent
neural networks.
However, how to extend the captioning techniques to describe videos
remains an open question.


Over the past several years, various methods have been proposed for
generating video descriptions.
Early efforts~\cite{yao2015describing} simply extend the encoder-decoder paradigm
in image captioning to videos.
Such methods follow a similar pipeline, namely
embedding an entire video into a feature vector,
feeding it to a decoding network to obtain a descriptive sentence.
However, for a video with rich temporal structures, a single sentence
is often not enough to capture all important aspects of the underlying
events.
Recently, a new stream of efforts emerge~\cite{krishna2017dense, shen2017weakly},
which attempt to use multiple sentences to cover a video.
Whereas such methods can provide more complete characterization
of a video, they are still subject to various issues,
\eg~lack of coherence among sentences and high redundancy.
These issues, to a large extent, are ascribed to two reasons:
(1) failing to align the temporal structure of the given video
with the narrative structure of the generated description; and
(2) neglecting the dependencies among sentences.


In this work, we aim to develop a new framework for generating
\emph{paragraph} descriptions for videos with rich temporal structures.
The goal is to generate descriptions that are
\emph{relevant}, \emph{coherent}, and \emph{concise}.
According to the discussion above,
the key to achieving this goal lies in two aspects:
(1) associate the temporal structures in the given video
with the linguistic generation process; and
(2) encourage coherence among sentences in an effective way.


Specifically, our approach is based on two key observations.
First, a natural video is usually composed of multiple short and meaningful
segments that reflect a certain step in a procedure or an episode of
a story. We refer to such video segments as \emph{events}.
While a single sentence may not be enough to describe a long video,
it often suffices to characterize an individual event.
Second, when people describe a video with a paragraph, there exist strong
logical and linguistic relations among consecutive sentences.
What to describe in a sentence depends strongly on what have been said.

Inspired by these observations, we devise a two-stage framework.
This framework first localizes candidate events from the video through
video action detection.
On top of these candidates, the framework then generate a coherent paragraph
in a \emph{progressive} manner.
At each step, it selects the next event to describe and produces a sentence therefor,
both conditioned on what have been said before.
The progressive generation process is driven by a variant of LSTM network
that takes into account both temporal and linguistic structures.
To effectively learn this network, we adopt the \emph{self-critical sequence training}
method and introduce rewards in two different levels, namely
the sentence level and the paragraph level.
%
%
On ActivityNet Captions~\cite{krishna2017dense}, the proposed framework
outperforms previous ones under multiple metrics.
Qualitatively, the descriptions produced by our method are
generally more relevant, more coherent, and more concise.


The key contribution of this work lies in a new framework for generating
descriptions for given videos. This framework is distinguished from previous
ones in three key aspects:
(1) It aligns the temporal structure of the given video and the narrative
structure of the generated description via a recurrent network.
(2) It maintains coherence among the sentences in a paragraph by explicitly
conditioning \emph{what to say next} on both the temporal structures and
\emph{what have been said}.
(3) It is learned via reinforcement learning, guided by rewards in both
the sentence level and the paragraph level.


\section{Related Work}
\label{sec:relwork}

\subsubsection{Image Captioning}
Early attempts of image captioning rely on visual concept detection,
followed by templates filling \cite{kulkarni2013babytalk} or nearest neighbour retrieving \cite{farhadi2010every}.
Recently, Vinyals~\etal \cite{vinyals2015show} proposed
the encoder-decoder paradigm,
which extracts image features using a CNN, followed by an RNN as the decoder to generate captions based on the features.
This model outperforms classical methods and becomes the backbone of state-of-the-art captioning models.
Many variants~\cite{xu2015show, dai2017towards, dai2017contrastive} are proposed following the encoder-decoder paradigm,
For example, Xu~\etal \cite{xu2015show} improved it by introducing an attention mechanism to guide the decoding process.

While many methods for image captioning can be seamlessly converted into methods for video captioning,
video contains richer semantic content that spreads on the temporal dimension,
directly applying methods for image captioning often lead to the loss of temporal information.

\subsubsection{Video Dense Captioning}
Video dense captioning, is a topic that closely related to video captioning,
where it densely generates multiple sentences,
covering different time spans of the input video.
Specifically,
Krishna \etal~\cite{krishna2017dense} proposed a method that obtains a series of proposals from the input video
and uses a captioning model to generate a sentence for each,
where the temporal relationships among the proposals are taken into account.
On the other hand, Shen \etal~\cite{shen2017weakly} proposed a weakly-supervised method,
which uses multi-instance multi-label learning to detect words from the input video
and then uses these words to select spatial regions to form region sequences.
Finally, it employs a sequence-to-sequence submodule to transfer region sequences into captions.

Although closely related,
video dense captioning is different from video captioning.
Particularly,
a model for video dense captioning could generate multiple captions,
each covers a small period of the input video,
where the periods can be overlapped with each other,
leading to a lot of redundancy in the corresponding captions.
On the contrary,
a model for video captioning should generate
a \emph{single} description consisting of several \emph{coherent} sentences
for the \emph{entire} input video.

\subsubsection{Video Captioning}
Our method targets the topic of video captioning.
Related works can be roughly divided into two categories based on whether a single sentence or a paragraph is generated for each input video.
In the first category,
a single sentence is generated.
Among all the works in this category,
Rohrbach \etal \cite{rohrbach2015long} detected a set of visual concepts at first, including verbs, objects and places,
and then applied an LSTM net to fuse these concepts into a caption.
Yu \etal \cite{yu2017end} and Pan \etal \cite{pan2016video} followed a similar way,
but respectively using a semantic attention model and a transfer unit to select detected concepts and generate a caption.
Instead of relying on visual concepts,
Hori \etal~\cite{hori2017attention} and Venugopalan \etal \cite{venugopalan2015sequence}
use features from multiple sources including appearance and motion to improve quality of the generated caption.
There are also efforts devoted to improving the decoder side.
Wang \etal \cite{wang2016multimodal} added a memory network before the LSTM net during the decoding process
to share features at different timestamps.
Baraldi \etal \cite{baraldi2016hierarchical} applied a boundary detecting module to share features hierarchically.
While they are able to produce great captions,
a single sentence is difficult to capture all semantic information in a video,
as one video usually contains several distintive events.

The second category is to generate a paragraph to describe a video.
Our method belongs to this category.
In this category, Yu \etal \cite{yu2016video} applied hierarchical recurrent neural networks,
where a sentence generator is used to generate a single sentence according to a specific \emph{topic},
and a paragraph generator is used to capture inter-sentence statistics and
feed the sentence generator with a series of topics.
The most similar work to our method is the one presented in~\cite{sah2017semantic}.
This method first select a subset of clips from the input video,
and then use a decoder to generate sentence from these clips to form a paragraph
summary of the entire video.
Our method is different from these existing works from two aspects.
(1) When generating each sentence of the paragraph, the method in \cite{yu2016video}
requires the features from the entire video, which is expensive for very long videos,
while our method only requires features in selected proposals.
(2) In \cite{sah2017semantic}, the clips are selected according to frame quality
\emph{in advance} as a preprocessing step, without taking into account the coherence of narration.
This way will lead to redundancy in the resulting paragraph.
On the contrary, our method selects key events \emph{along with} the generation of captions
in a progressive way. The selection of the next key event depends on
what has been said before in preceding captions.
Also, this process takes into account temporal and semantic relationships among
the selected events in order to ensure the coherence of the resultant paragraph.


\begin{figure*}
\centering
\includegraphics[width=0.95\textwidth]{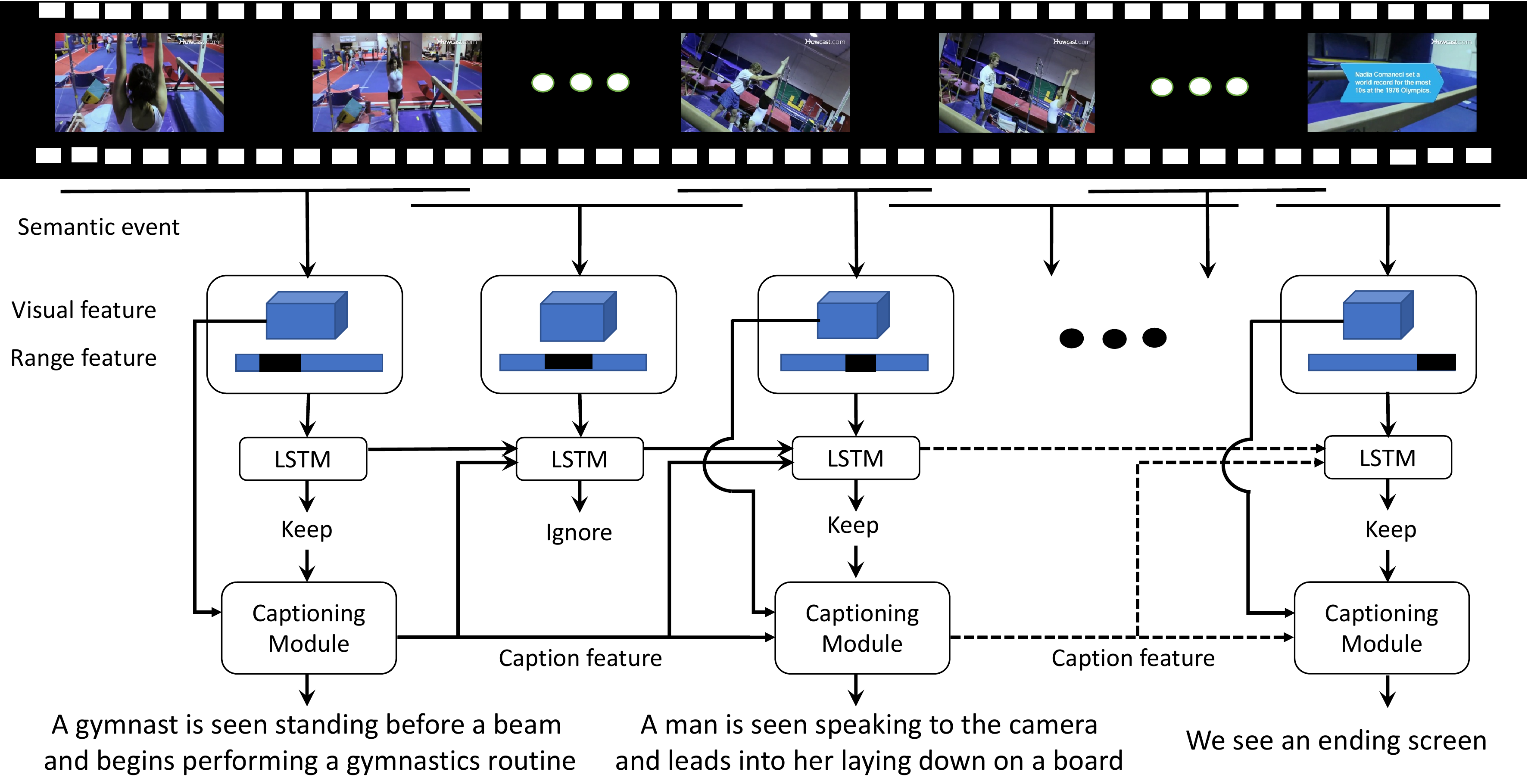}
\caption{
   An overview of our framework, which at first localize important events from the entire video.
It then generates a coherent and concise descriptive paragraph upon these localized events.
Specifically, an LSTM net, serves as a selection module, will pick out a sequence of coherent and semantically independent events,
based on appearances, temporal locations of events, as well as their semantic relationships.
And based on this selected sequence, another LSTM net, serves as a captioning module,
will generate a single sentence for each event in the sequence conditioned on previous generated sentences,
which are then concatenated sequentially as the output of our framework.
}
\label{fig:overview}
\end{figure*}

\section{Generation Framework}
\label{sec:method}

Our task is to develop a framework that can generate coherent
paragraphs to describe given videos.
Specifically, a good description should possess three properties:
(1) \emph{Relevant}: the narration aligns well with the events
in their temporal order.
(2) \emph{Coherent}: the sentences are organized into a logical and
fluent narrative flow.
(3) \emph{Concise}: each sentence conveys a distinctive message, without
repeating what has been said.

\subsection{Overview}

A natural video often comprises multiple events that are located
sparsely along the temporal range.
Here, \emph{events} refer to those video segments that contain distinctive
semantics that need to be conveyed.
Taking the entire video as input for generating the description is inefficient,
and is likely to obscure key messages when facing numerous noisy clips.
Therefore, we propose a framework as shown in Figure \ref{fig:overview},
which generates a descriptive paragraph in two stages,
namely \emph{event localization} and \emph{paragraph generation}.
In event localization,
we localize \emph{candidate events} in the video with a high recall.
In paragraph generation,
we first filter out redundant or trivial candidates,
so as to get a sequence of important and distinctive events.
We then use this sequence to generate a single descriptive paragraph
for the entire video in a progressive manner, taking into account
the coherence among sentences.

\subsection{Event Localization}

To localize event candidates, we adopted the clip proposal generation
scheme presented in~\cite{SSN2017ICCV}, using the released codes.
This scheme was shown to be effective in locating important clips
from untrimmed videos with high accuracy.
Specifically, following~\cite{SSN2017ICCV},
we calculate frame-wise importance scores,
and then group the frames into clips of via a watershed procedure.
This method outputs a collection of clips as \emph{event candidates}.
These clips have varying durations and can overlap with each other.
As our focus is on paragraph generation,
we refer readers to \cite{SSN2017ICCV} for more details of event localization.

Note that not all the candidates derived
in this stage is worthy of description. The paragraph generation
will select a subset of candidates that
contain important and distinctive messages, \emph{along with}
the generation process.

\subsection{Progressive Event Selection and Captioning}

Given a sequence of events, there are various ways to generate a descriptive
paragraph. While the most straightforward way is to generate a single sentence
for each event in the sequence, it is very likely to introduce a large amount of
redundancy.
To generate coherent and concise description, we can select a subset of
distinctive events and generate sentences thereon. The key challenge here
is to stride a good balance between sufficient coverage and conciseness.

In this work, we develop a \emph{progressive} generation framework that
couples two recurrent networks, one for event selection and the other for
caption generation.

\subsubsection{Event Selection}
With all event candidates arranged in the chronological order,
denoted as $(e_1, \ldots, e_T)$, the event selection network
begins with the first candidate in the sequence and moves forward gradually
as follows:
\begin{align}
	\vh_0 & = \vzero, \ \
	\vh_t = \text{LSTM}(\vh_{t - 1}, \vv_t, \vr_t, \vc_{k_t}), \\
	p_t & = \text{sigmoid}(\vw_p^T \vh_t), \ \
	y_t = 1[p_t > \delta].
\end{align}
Specifically, it initializes the latent state vector $\vh_0$ to be zeros.
At each step $t$, it updates the latent state $\vh_t$ with an LSTM cell
and computes $p_t$, the probability of $e_t$ containing relevant and
distinctive information, by applying a sigmoid function to $\vw_p^T \vh_t$.
If $p_t$ is above a threshold $\delta$, $y_t$ would be set to $1$, indicating
that the candidate $e_t$ will be selected for sentence generation.

The updating of $\vh_t$ depends on four different inputs:
(1) $\vh_{t-1}$: the latent state at the preceding step.
(2) $\vv_t$: the visual feature of $e_t$, extracted using
a Temporal Segmental Network (TSN)~\cite{wang2016temporal}.
(3) $\vr_t$: the range feature,
similar as the image mask in~\cite{dai2017detecting},
represented by a binary mask that indicates the normalized time span of $e_t$
relative to the entire duration.
(4) $\vc_{k_t}$: the caption feature of $e_{k_t}$, where $k_t$ is the index of
the \emph{last} selected event candidate (before $t$).
Here, the caption feature is from the \emph{caption generation network},
which we will present next. Particularly, $\vc_{k_t}$ is chosen to be
the latent state of the caption generation network at the final decoding step
when generating the description for $e_{k_t}$.

With the previous caption feature $\vc_{k_t}$ incorporated, the event selection
network is aware of what have been said in the past when making a selection.
This allows it to avoid selecting the candidates that are semantically redundant.

\subsubsection{Caption Generation}
On top of the selected events, the \emph{caption generation network}
will produce a sequence of sentences, one for each event, as follows:
\begin{align}
	\vg_0^{(k)} & = \begin{cases} \vzero, & k = 1 \\
					\vg_{*}^{(k - 1)}, & k > 1
			\end{cases}, \quad
	\vg_l^{(k)} = \text{LSTM}(\vg_{l-1}^{(k)}, \vu_l^{(k)}, w_{l-1}^{(k)}), \\
	\vs_l^{(k)} &= \mW_{s} \vg_l^{(k)}, \quad
	w_l^{(k)} \sim \text{softmax}(\vs_l^{(k)}).
\end{align}
Here, $\vg_l^{(k)}$ denotes the latent state at the $l$-th step of
the caption generation network when describing the $k$-th selected event.
$\vu_l^{(k)}$ denotes a visual feature of a subregion of the event.
Here, the computation of $\vu_l^{(k)}$ follows the scheme presented
in~\cite{rennie2016self}, which allows the network to dynamically attend to different
subregions as it proceeds\footnote{We provide more details of the computation
of $\vu_l^{(k)}$ in the supplemental materials}.
$w_l^{(k)}$ is the word produced at the $l$-th step, which is sampled from
$\text{softmax}(\vs_l^{(k)})$.

This network is similar to a standard LSTM for image captioning except for
an important difference:
When $k > 1$, the latent state is initialized to $g_*^{(k-1)}$, the latent state
at the last decoding step while generating the previous sentence.
This means that the generation of each sentence (except the first one) is
conditioned on the preceding one,
which allows the generation to take into account the coherence among sentences.

\subsubsection{Discussions}
The event selection network and the caption generation network
work hand in hand with each other when generating a description for a given video.
On one hand, the selection of next event candidate depends on what has been said.
Particularly, one input to the event selection network is $\vc_{k_t}$, which is set to
the $\vg_*^{(k_t)}$, the last latent state of the caption generation network in
generating the previous sentence.
On the other hand, the caption generation network is invoked only when the event
selection network outputs $y_t = 1$, and the generation of the current sentence
depends on those that come before.
The collaboration between both networks allows the framework to produce paragraphs
that can cover major messages, while being coherent and concise.

Note that
one may also use Non-Maximum Suppression (NMS) to directly remove temporal overlapped events.
This simple way is limited compared to ours, as it only considers temporal overlap
while ignoring the semantic relevance.
Another way is to first generate sentences for all events and then select a subset
of important ones based on text summarization~\cite{sah2017semantic}.
This approach, however, does not provide a mechanism to encourage linguistic
coherence among sentences, which is crucial for generating high-quality descriptions.


\section{Training}
\label{sec:train}

Three modules in our framework need to be trained, namely
\emph{event localization},
\emph{caption generation}, and
\emph{event selection}.
In particular,
we train the event localization module simply following the procedure
presented in~\cite{SSN2017ICCV}.
The other two modules, the caption generation network and the event selection
network, are trained separately.
We first train the caption generation network using the ground-truth event captions.
Thereon, we then train the event selection network, which requires the
caption generation states as input.

\subsection{Training Caption Generation Network}

The caption generation network models the distribution of each word conditioned
on previous ones and other inputs, including the visual features of the
corresponding event $\vu^{(k)}$ and the final latent state of for the preceding
sentence $\vg_*^{(k-1)}$.
Hence, this distribution can be expressed as
$p_{\vtheta}(w_l | w_{1:l-1}; \vu^{(k)}, \vg_*^{(k-1)})$,
where $\vtheta$ denotes the network parameters.
We train this network through two stages: (1) initial supervised training,
and (2) reinforcement learning.

The initial supervised training is performed based on pairs of
events and their corresponding ground-truth descriptions, with the
standard cross entropy loss.
Note that this network requires $\vg_*^{(k-1)}$ as input, which is provided
\emph{on the fly} during training. In particular, we feed the ground-truth
sentences for each video one by one. At each iteration, we cache the final
latent state for the current sentence and use it as an input for the next one.

Supervised training encourages the caption generation network to
emulate the training sentences word by word. To further improve the quality
of the resultant sentences, we resort to reinforcement learning.
In this work, we employ the \emph{Self-Critical Sequence Training (SCST)}~\cite{rennie2016self}
technique.
Particularly, we consider the caption generation network as an \emph{``agent''},
and choosing a word an \emph{``action''}.
Following the practice in~\cite{rennie2016self}, we update the network
parameters using approximated policy gradient in the reinforcement learning stage.

The key to reinforcement learning is the design of the rewards.
In our design, we provide \emph{rewards} at two levels, namely
the \emph{sentence-level} and the \emph{paragraph-level}.
As mentioned, the network takes in the ground-truth events of a video sequentially,
producing one sentence for each event (conditioned on the previous state).
These sentences together form a paragraph.
When a sentence is generated, it receives a sentence-level reward.
When the entire paragraph is completed, it receives a paragraph-level reward.
The reward is defined to be the CIDEr~\cite{vedantam2015cider} metric between the
generated sentence/paragraph and the ground truth.

\subsection{Training Event Selection Network}

The event selection network is a recurrent network that takes a sequence of
candidate events as input and produces a sequence of binary indicators
(for selecting a subset of candidates to retain).
We train this network in a supervised manner. Here, the key question is how
to obtain the training samples.
We accomplish this in two steps: (1) labeling and (2) generating training
sequences.

First, for each video, we use the event localization module to produce a series
of event candidates as $(e_1, \ldots, e_T)$.
At the same time, we also have a set of ground-truth events provided by the
training set, denoted as $(e^*_1, \ldots, e^*_{T^*})$.
For each ground-truth event $e^*_j$, we find the candidate $e_i$ that has
the highest overlap with it, in terms of the temporal IoU, and label it
as positive, \ie~setting $y_i = 1$.
All the other event candidates are labeled as negative.

Second, to generate training sequences, we consider three different ways:

\begin{itemize}
\item \emph{(S1)} \emph{Complete sequences}, which simply uses the whole sequence
of candidates for every video, \ie~$(e_1, e_2, \cdots, e_m)$.
\item \emph{(S2)} \emph{Subsampling at intervals},
which samples event candidates at varying intervals, \eg~$(e_2, e_4, \cdots, e_m)$,
to obtain a larger set of sequences.
\item \emph{(S3)} \emph{Subsampling negatives}, which keeps all positive candidates,
while randomly sampling the same number of negative candidates in between.
\end{itemize}

Note that the positive and negative candidates are highly imbalanced.
For each video, positive candidates are sparsely located, while negative ones
are abundant.
The scheme (S3) explicitly rebalances their numbers.
Our experiment shows that (S3) often yields the best performance.

The event selection network is trained with the help of the
caption generation network.
To be more specific, whenever the event selection network yields
a positive prediction, the caption feature $\vc_{k_t}$ will be updated
based on the caption generation network, which will be fed as an input
to the next recurrent step.


\section{Experiment}
\label{sec:exp}

We report our experiments on ActivityNet Captions \cite{krishna2017dense},
where we compared the proposed framework to various baselines,
and conducted ablation studies to investigate its characteristics.

\subsection{Experiment Settings}

The ActivityNet Captions dataset \cite{krishna2017dense}
is the largest publicly available dataset for video captioning.
This dataset contains
$10,009$ videos for training, and $4,917$ for validation.
Compared to previous datasets~\cite{geiger2013vision, regneri2013grounding, rohrbach2014coherent},
it has two orders of magnitude more videos.
The videos in this dataset are $3$ minutes long on average.
Each video in the training set has one set of human labeled annotations,
and each video in the validation set has two such sets.
Here,
a set of annotations is a series of sentences,
each aligned with a long or short segment in the video.
About $10\%$ of all segments have overlaps with each other.
On average, each set of annotations contains $3.65$ sentences.

While ActivityNet Captions was originally designed for the task of video dense captioning,
we adapt it to our task with the ground-truth paragraphs derived by
sequentially concatenating the sentences within each set of segment-based annotations.
As a result, for each video in the training set, there is one ground-truth paragraph,
and for each video in the validation set, there are two ground-truth paragraphs.
Since the annotations for testing videos are not publicly accessible,
we randomly split the validation set in half,
resulting in $2,458$ videos for tuning hyperparameters,
and $2,459$ videos for performance evaluation.

We set $\delta = 0.3$, $L = 100$ in the event selection module,
and $N = 10$ in the captioning module.
We separately train the three modules in our framework.
In particular, 
the event localization module is trained according to \cite{SSN2017ICCV}, 
where the localized events have a recall of 63.77\% at 0.7 tIoU threshold. 
For captioning module, the LSTM hidden size is fixed to 512. 
As discussed in section~\ref{sec:train}, we first train the model under the cross-entropy objective using ADAM~\cite{kingma2014adam} with an initial learning rate of $4 \times 10^{-4}$. 
We choose the model with best CIDEr score on the validation set.
We then run SCST training initialized with this model.
The reward metric is CIDEr, for both sentence and paragraph reward. 
SCST training uses ADAM with a learning rate $5 \times 10^{-5}$. 
One batch contains at least $80$ events, since events in the same video are fed into a batch simultaneously. 
For the event selection module,
we train it on a collection of labeled training sequences prepared as described in
Section~\ref{sec:train}, where each training sequence contains $64$ candidate events.
We use cross-entropy as the loss function
and SGD with momentum as the optimizer.
The learning rate is initialized to $0.1$ and scaled down by a factor of $0.1$
every $10,000$ iterations.
We set the SGD momentum to $0.9$, weight decay to $0.0005$, and batch size to $80$. \footnote{Code will be made publicly available soon}

\subsection{Evaluation}
We evaluate the performance
using multiple metrics, including BLEU~\cite{papineni2002bleu}, METEOR~\cite{denkowski2014meteor},
CIDEr~\cite{vedantam2015cider} and Rouge-L~\cite{lin2004rouge}.
Besides, we notice there exists a general problem for video captioning results,~\ie, repetition or redundancy.
This may be due to that captioning module can't distinguish detailed events. 
For example, the captioner may take both ironing collar and ironing sleeve as ironing. 
Describing repeated things definitely hurts the coherence of the descriptions.
However this can not reflect in the above metrics. 

To measure this effect, a recent work proposes Self-BLEU~\cite{zhu2018texygen} by
evaluating how one sentence resembles the rest a generated paragraph. 
We also propose another metric, called Repetition Evaluation (RE).
Given a generating description $c_i$, the number of time an $n$-gram $w_k$ occurs in it is denoted as $h_k(c_i)$. 
The Repetition Evaluation computes a redundancy score for every description $c_i$:
\begin{align}
    RE(c_i) = \frac{\sum_k{\max{(h_k(c_i) - 1, 0)}}}{\sum_k{h_k(c_i)}},
\end{align}
where the gram length, $n$, takes a large number, like 4 in our experiments. 
The corpus-level score is the mean score across all descriptions. 
Ideally, a description with $n$ repetitions would get a score around $(n-1)/n$.

\subsection{Comparison with Other Methods}

We compared our framework with various baselines,
which we describe as below.

(1) \emph{Sentence-Concat}:
a simple baseline that equally splits a video into four disjoint parts,
and describes each part with a single sentence using the captioning model as ours. 
The final paragraph is derived by concatenating these four sentences.
Using this baseline, we are able to study the effect of localizing events in the input video.

(2) \emph{Hierarchical-RNN} \cite{yu2016video}:
a more sophisticated way to generate paragraphs from a video,
where a topic RNN generates a sequence of topics to control the generation
of each individual sentence. With the topic embeddings as input,
a sentence RNN generates a sentence for each topic in the sequence.

(3) \emph{Dense-Caption} \cite{krishna2017dense}:
one of the state-of-the-art methods for video dense captioning,
which generates a single sentence for each candidate event
and then concatenate them all into a paragraph,
regardless of their similarities.
This baseline is used to demonstrate the differences between
\emph{video dense captioning} and \emph{video captioning}.

(4) \emph{Dense-Caption-NMS}:
a method based on the above Dense-Caption. 
It selects events from candidate events of Dense-Caption using Non-Maximum Suppression (NMS),
removing those that are highly overlapped with others on temporal range. 

(5) \emph{Semantic-Sum} \cite{sah2017semantic}:
a recent method that also identifies the video segments as ours.
We find that this method gets best performance when setting sentence length as 3 and using Latent Semantic Analysis~\cite{steinberger2004using} in summarization module. 

(6) \emph{Move Forward and Tell (MFT)}:
our proposed framework which progressively select events and produce sentences conditioned on what have been said before. 

(7) \emph{GT-Event}:
this baseline directly applies our captioning module on \emph{ground-truth events}.
In principle, this should serve as a performance upper bound, as it has access
to the ground-truth event locations.

\begin{table*}[t]
\centering
\caption{\footnotesize
    Comparison results of video captioning. The last method GT-Event uses ground truth information and can be seen as the upper bound of event selecting. The value is in \%.
For metrics RE and Self-BLEU, the lower is better. For the others, the higher is better}
\resizebox{\columnwidth}{!}{%
\begin{tabular}{lccccccccc}
\toprule
Model             & CIDEr          & BLEU@4        & BLEU@3         & BLEU@2         & BLEU@1         & Rouge-L       & METEOR         & RE             & Self-BLEU      \\
\midrule
Sentence-Concat   & 4.51           & 4.18          & 6.45           & 10.41          & 17.52          & 22.59          & 8.79           & 18.41          & 49.40          \\
Hierarchical-RNN \cite{yu2016video}  & 6.99           & 7.32          & 12.13          & 21.23          & 39.02          & 25.53          & 10.79          & 18.79          & 54.38          \\
Dense-Caption \cite{krishna2017dense}     & 0.29           & 0.99          & 1.51           & 2.33           & 3.52           & 8.63           & 6.75           & 64.05          & 89.79          \\
Dense-Caption-NMS & 3.52           & 4.45          & 6.96           & 11.21          & 18.09          & 21.41          & 12.08          & 23.98          & 62.46          \\
Semantic-Sum \cite{sah2017semantic}      & 10.43          & 6.44          & 11.21          & 20.36          & 37.22          & 25.44          & 12.66          & 29.94          & 67.49          \\
MFT (Ours)        & \textbf{14.15} & \textbf{8.45} & \textbf{13.52} & \textbf{22.26} & \textbf{39.11} & \textbf{25.88} & \textbf{14.75} & \textbf{17.59} & \textbf{45.80} \\
\midrule
GT-Event          & 19.56          & 10.33         & 16.44          & 27.24          & 46.77          & 29.70          & 15.09          & 15.88          & 42.95           \\
\bottomrule
\end{tabular}
}
\label{tab:overall}
\end{table*}

Table~\ref{tab:overall} presents the results for different methods on ActivityNet Captions,
from which we have the following observations:
(1) \emph{Dense-Caption} performs very poorly since there are many redundant proposals and therefore repeated sentences indicating by RE and Self-BLEU metrics.
The RE score is terribly high, meaning two thirds of the descriptions are likely redundant. 
This clearly shows the differences between video dense captioning and video captioning.
(2) \emph{Sentence-Concat} and \emph{Dense-Caption-NMS} are at a comparable level, much better than \emph{Dense-Caption}.
These two methods may benefit from a common aspect,~\ie their events are almost not overlapped.
But some important events may not be localized here.
(3) \emph{Hierarchical-RNN} improves the result to the next level, achieving $25.53\%$ Rouge-L.
This suggests that it produces more coherent result by modelling the sentence relationships. 
(4) \emph{Semantic-Sum} further improves the result, achieving $10.43\%$ CIDEr. 
This shows the effect of localizing video events for captioning. 
(5) Our method \emph{MFT} significantly outperforms all above methods,
\eg~it attains $14.15\%$ for CIDEr, compared to $10.43\%$ from \emph{Semantic-Sum}.
It also performs comparably well with \emph{GT-Event}, a method utilizing ground-truth events. 


\subsection{Ablation Studies}

\subsubsection{Training Scheme}

In section~\ref{sec:train}, we introduce our training strategies. 
For caption generation network, the training includes two phases -- (P1) \emph{initial supervised training} and (P2) \emph{reinforcement learning}. 
For event selection network, we propose three different ways to generate training sequences. 
(S1) \emph{complete sequences},
(S2) \emph{subsampling at intervals} and
(S3) \emph{subsampling negatives}
where our model adopts scheme (S3). 

\begin{table*}[t]
\centering
\caption{\footnotesize
    This table compares different training schemes for our model. See section~\ref{sec:train} for the detail of schemes}
\resizebox{\columnwidth}{!}{%
\begin{tabular}{lccccccccc}
\toprule
Training Scheme               & CIDEr          & BLEU@4        & BLEU@3         & BLEU@2         & BLEU@1         & Rouge-L        & METEOR         & RE             & Self-BLEU      \\
\midrule
(S1) Complete sequenses       & 9.25           & 7.12          & 11.56          & 19.18          & 33.80          & 23.84          & 11.06          & 26.11          & 51.41          \\
(S2) Subsampling at intervals & 8.04           & 6.49          & 10.52          & 17.58          & 31.16          & 21.27          & 10.16          & 24.09          & 46.19          \\
(P1) Supervised training only & 12.81          & \textbf{8.53} & 13.25          & 21.76          & 37.68          & 25.53          & 13.24          & 19.10          & 46.77          \\
(P1 + P2 \& S3) Ours          & \textbf{14.15} & 8.45          & \textbf{13.52} & \textbf{22.26} & \textbf{39.11} & \textbf{25.88} & \textbf{14.75} & \textbf{17.59} & \textbf{45.80} \\
\bottomrule
\end{tabular}
}
\label{tab:train}
\end{table*}

Table~\ref{tab:train} shows the performance under different training schemes. 
First of all, among all three schemes of generating training sequences for event selection network, (S3) \emph{subsampling negatives} performs best. 
(S1) scheme only gets $9.25\%$ CIDEr while (S2) scheme gets $8.04\%$ CIDEr, a considerable drop from our model. 
This indicates the importance of balancing the positive candidates and negative candidates. 
Besides, by utilizing the \emph{reinforcement learning} for caption generation network, 
our model gains performance improvement, especially for the target metric, CIDEr.
Our model yields $14.15\%$ CIDEr score while only using \emph{supervised training} yields $12.81\%$. 
Other metrics get consistent improvement with reinforcement learning, including RE and Self-BLEU metrics. 

\subsubsection{Feature for Event Selection}

In this part, we study the effect of using different features for event selection.
Specifically, the following combinations are tested.
(1) \emph{visual}: using visual features $\vv$ in isolation,
which serves as a baseline for other combinations.
(2) \emph{visual + range}: combining visual features $\vv$ with range features $\vt$,
which provides the temporal ranges in the original videos in addition to visual features.
(3) \emph{visual + caption}: combining visual features $\vv$ with caption feature $\vc$,
which provides event selection module the ability to know \emph{what have been said}.
(4) \emph{visual + range + caption}: using all features.

\begin{table*}[t]
\centering
\caption{\footnotesize
    This table lists the results of using different feature combinations for event selection, where the full combination is shown to be the best configuration}
\resizebox{\columnwidth}{!}{%
\begin{tabular}{lccccccccc}
\toprule
Feature                  & CIDEr          & BLEU@4        & BLEU@3         & BLEU@2         & BLEU@1         & Rouge-L        & METEOR         & RE             & Self-BLEU      \\
\midrule
Visual                   & 10.19          & 6.41          & 10.41          & 17.50          & 31.47          & 25.74          & 12.05          & 29.00          & 56.57          \\
Visual + range           & 10.54          & 7.53          & 12.16          & 20.32          & 36.58          & 25.42          & 11.86          & 27.62          & 54.11          \\
Visual + caption         & 11.29          & 7.90          & 12.61          & 20.83          & 36.55          & 23.82          & 13.24          & 18.92          & 49.14          \\
Visual + range + caption & \textbf{14.15} & \textbf{8.45} & \textbf{13.52} & \textbf{22.26} & \textbf{39.11} & \textbf{25.88} & \textbf{14.75} & \textbf{17.59} & \textbf{45.80} \\
\bottomrule
\end{tabular}
}
\label{tab:feature}
\end{table*}

As shown in Table \ref{tab:feature},
compared to \emph{visual}, both \emph{visual + range} and \emph{visual + caption} outperforms it by a significant margin,
indicating both the temporal range features and the caption features are complementary to the visual information.
It is reasonable as the visual features contain no temporal information,
which is explicitly captured in temporal range features,
and semantically captured in caption features.
Moreover, caption features and temporal range features are also complementary to each other,
as the full combination \emph{visual + range + caption} outperforms both \emph{visual + range} and \emph{visual + caption}.

\subsection{Human Study}

\begin{figure}
\centering
\includegraphics[width=\textwidth]{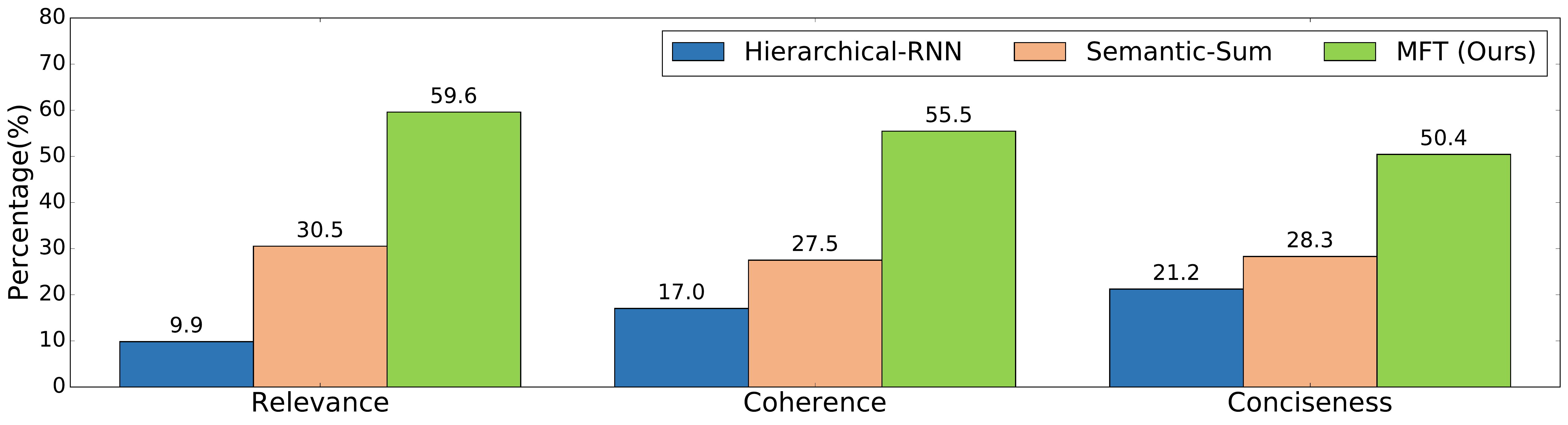}
\caption{
The figure shows the results of human evaluation, which compares paragraphs generated by different methods, with respect to relevance, coherence and conciseness.
}
\label{fig:human_study}
\end{figure}

To have a more pertinent assessment, we also conduct a human study,
where $20$ users are asked to pick the best descriptive paragraph among
those generated by the proposed MFT, Hierarchical-RNN, and Semantic-Sum,
in terms of relevance, coherence and conciseness respectively.
Figure \ref{fig:human_study} shows that MFT generates the best ones across all aspects.

\subsection{Qualitative Example}

\begin{figure*}
\centering
\includegraphics[width=\textwidth]{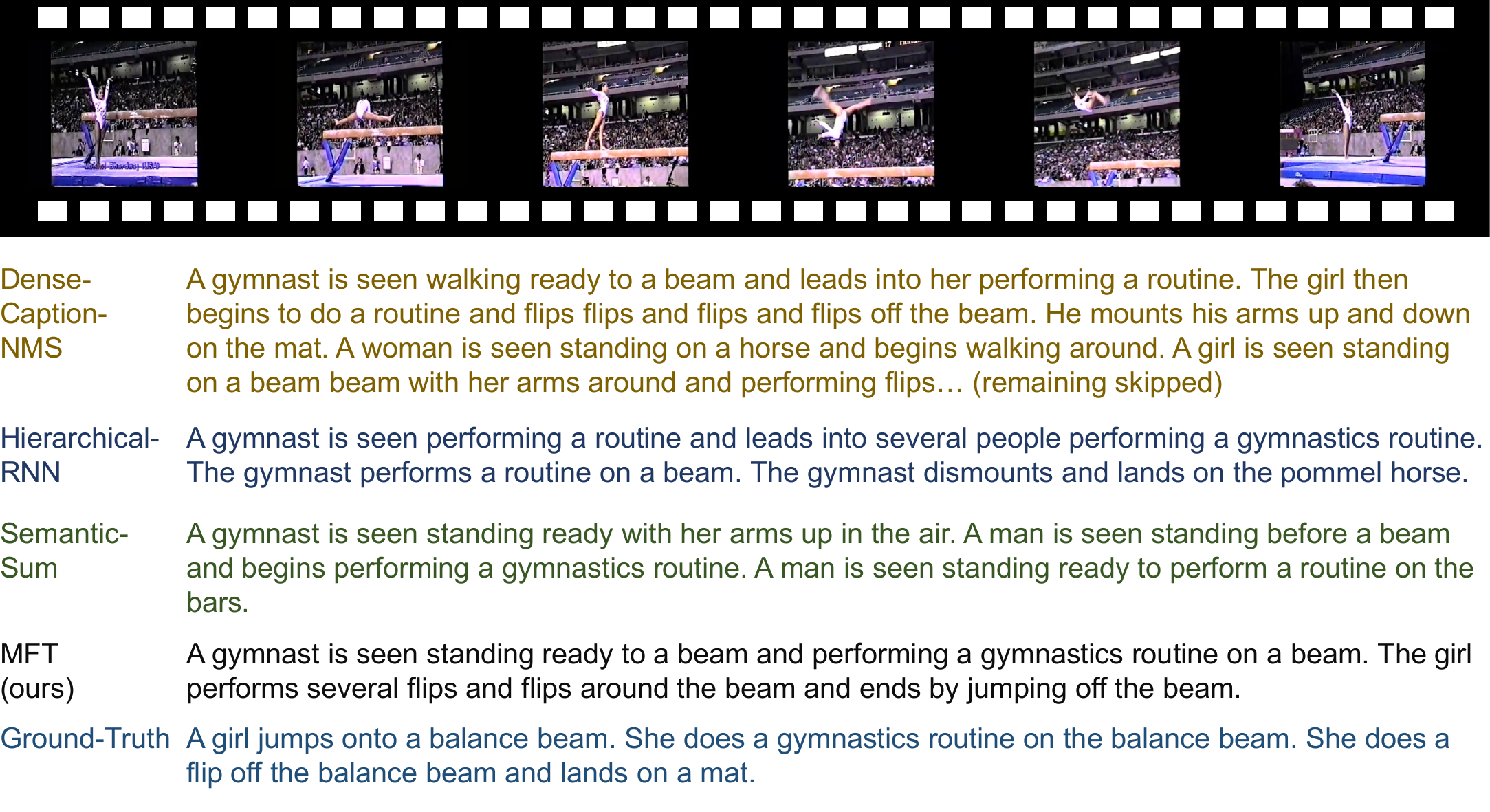}
\caption{
This figure lists a qualitative example, where paragraphs generated by \emph{Dense-Caption-NMS}, \emph{Hierarchical-RNN}, \emph{Semantic-Sum} and our method \emph{MFT} are shown.} 
\label{fig:sample}
\end{figure*}

In Figure \ref{fig:sample}, we also include one qualitative example,
where a video is shown with paragraphs generated by \emph{Dense-Caption-NMS}, \emph{Hierarchical-RNN}, \emph{Semantic-Sum} and our method \emph{MFT}.
As shown in Figure \ref{fig:sample},
compared to the baselines, our method produces more concise and more coherent paragraphs.
We will provide more qualitative results in the supplementary materials.

\section{Conclusion}
\label{sec:concls}

We presented a new framework for generating coherent descriptions for given videos.
The generated paragraphs well align with the temporal structure of the given video,
covering major semantics without redundancy.
Specifically,
it sequentially locates important events via an LSTM net,
which selects events from a candidate pool,
based on their appearances, temporal locations and mutual semantic relationships.
When respectively describing events in the acquired sequence,
it explicitly decides \emph{what to say next} based on temporal structure and \emph{what have been said}.
On ActivityNet Captions, our method significantly outperformed others
on a wide range of metrics,
while producing more coherent and more concise paragraphs.

\subsubsection*{Acknowledgement}
This work is partially supported by the Big Data Collaboration Research grant from SenseTime Group (CUHK Agreement No. TS1610626), the Early Career Scheme (ECS) of Hong Kong (No. 24204215).

\bibliographystyle{splncs04}
\bibliography{ref}

\clearpage

\section*{Appendix}

\renewcommand*{\thesection}{\Alph{section}}
\setcounter{section}{0}

\section{Caption generation process}
As described in section 3.3, the \emph{caption generation network} produces a sequence of sentences, one for each event. 
As shown in the equation (3), the caption module updates the hidden state at each time step depends on a dynamically computed visual feature $\vu$.
Specifically, for describing the $k$-th selected event at the $l$-th step, the computation of $\vu_l^{(k)}$ is as following:
\begin{align}
	\alpha_{l, i} & = \vw_{\alpha}^T \cdot \text{tanh}(\mW_{\alpha d} \vd^{(k)}_i + \mW_{\alpha h} \vh_{l - 1}^{(k)}), \\
	\vu_l^{(k)} & = \sum_{i = 1}^N \alpha_{l, i} \vd^{(k)}_i, 
\end{align} 
where we divide the $k$-th selected event into $N$ subsegments, 
and take average of the frame features in each segment to get 
$\{\vd^{(k)}_1, \vd^{(k)}_2, ..., \vd^{(k)}_N\}$.
The visual feature here allows the network to dynamically attend to different subregions as it proceeds. 

\section{Qualitative samples}

We include five qualitative samples here. For each sample, we compare the results obtained from different methods, including Dense-Caption-NMS, Hierarchical-RNN, Semantic-Sum, and MFT (our proposed method), as well as the human provided ground-truth. From the results, we can see that the descriptions produced by our method are generally more coherent and more concise.

\begin{figure}
\centering
\begin{minipage}{\textwidth}
\centering
\includegraphics[width=\textwidth]{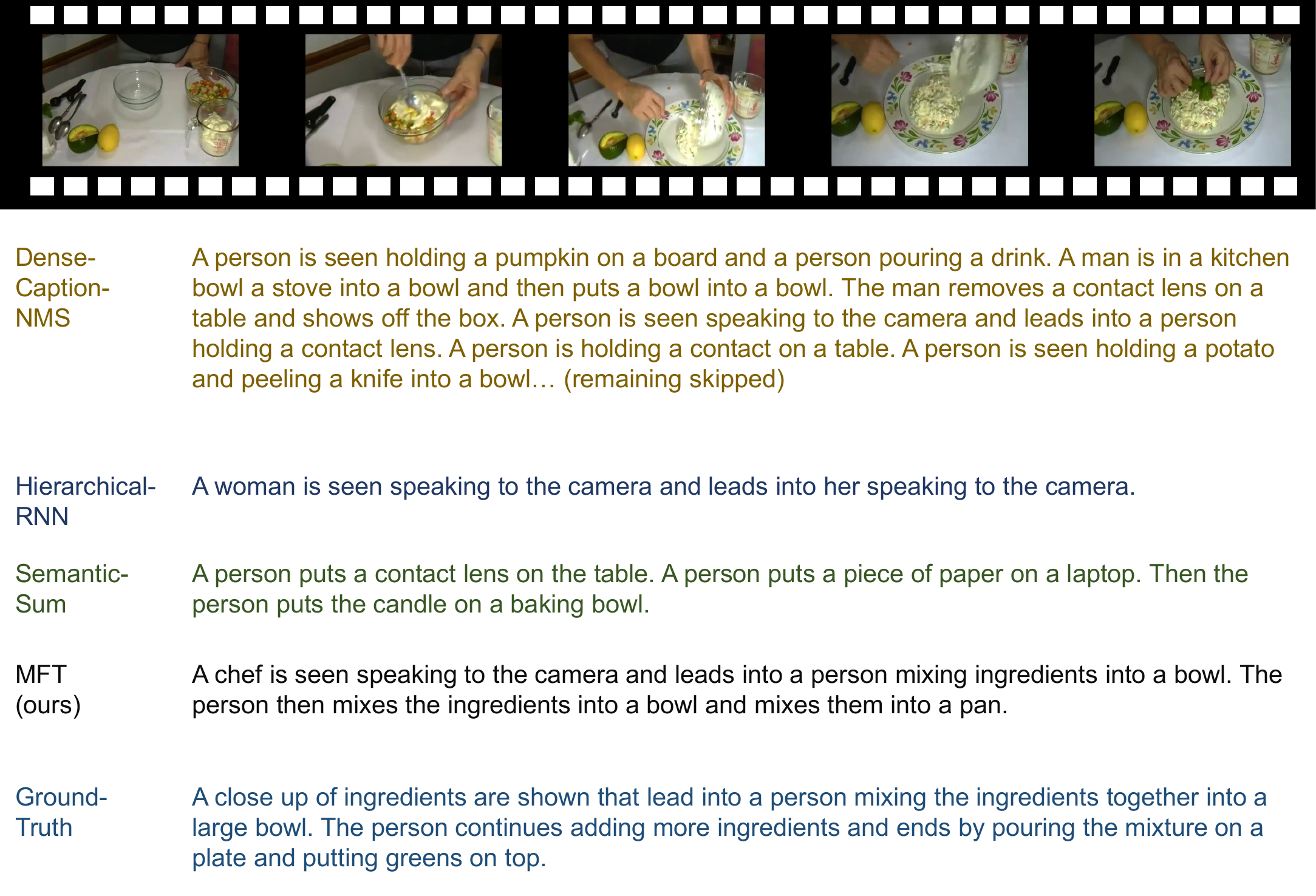}
\label{fig:sup1}
\end{minipage}

\begin{minipage}{\textwidth}
\centering
\includegraphics[width=\textwidth]{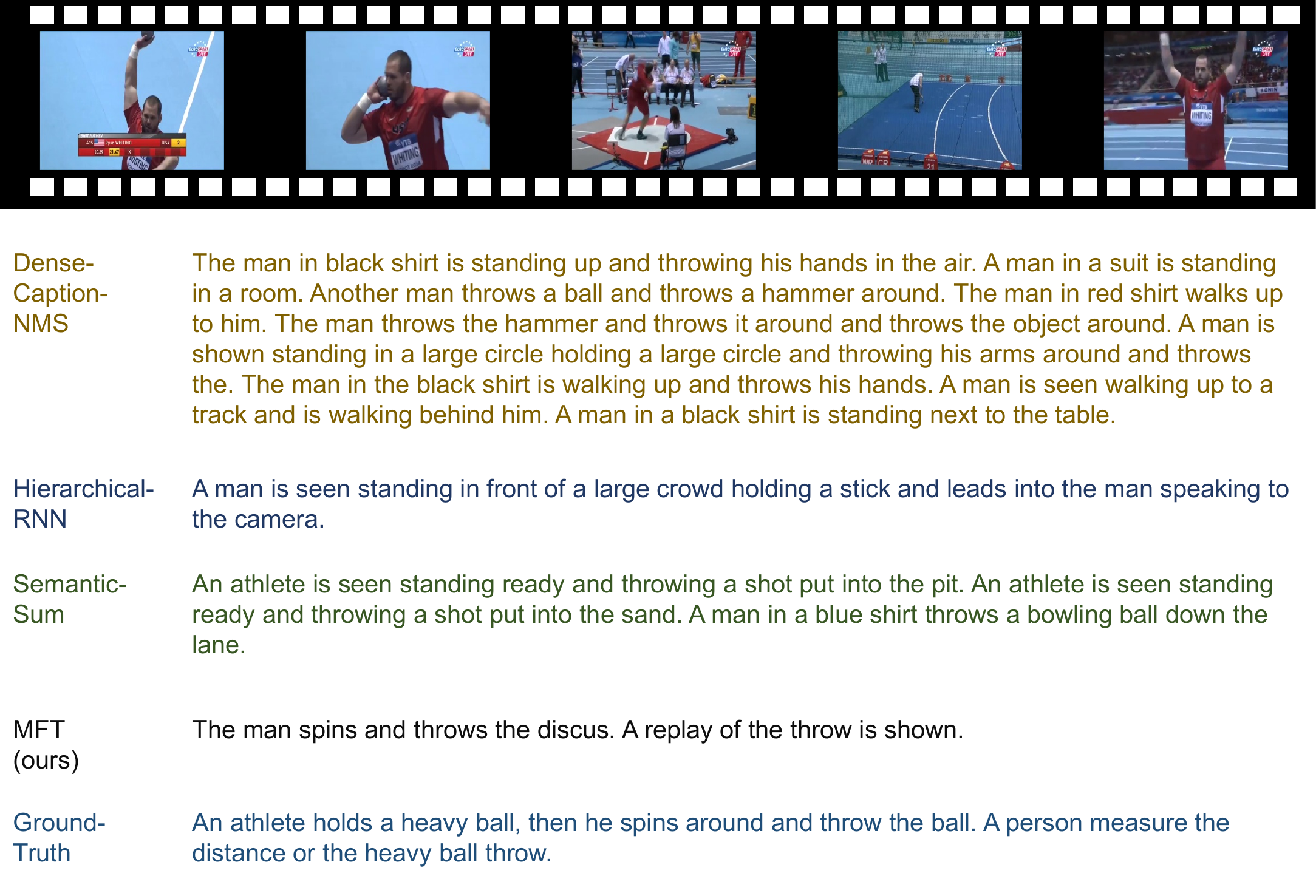}
\label{fig:sup2}
\end{minipage}
\caption{The first two samples. }
\end{figure}

\clearpage

\begin{figure}
\centering
\begin{minipage}{\textwidth}
\centering
\includegraphics[width=\textwidth]{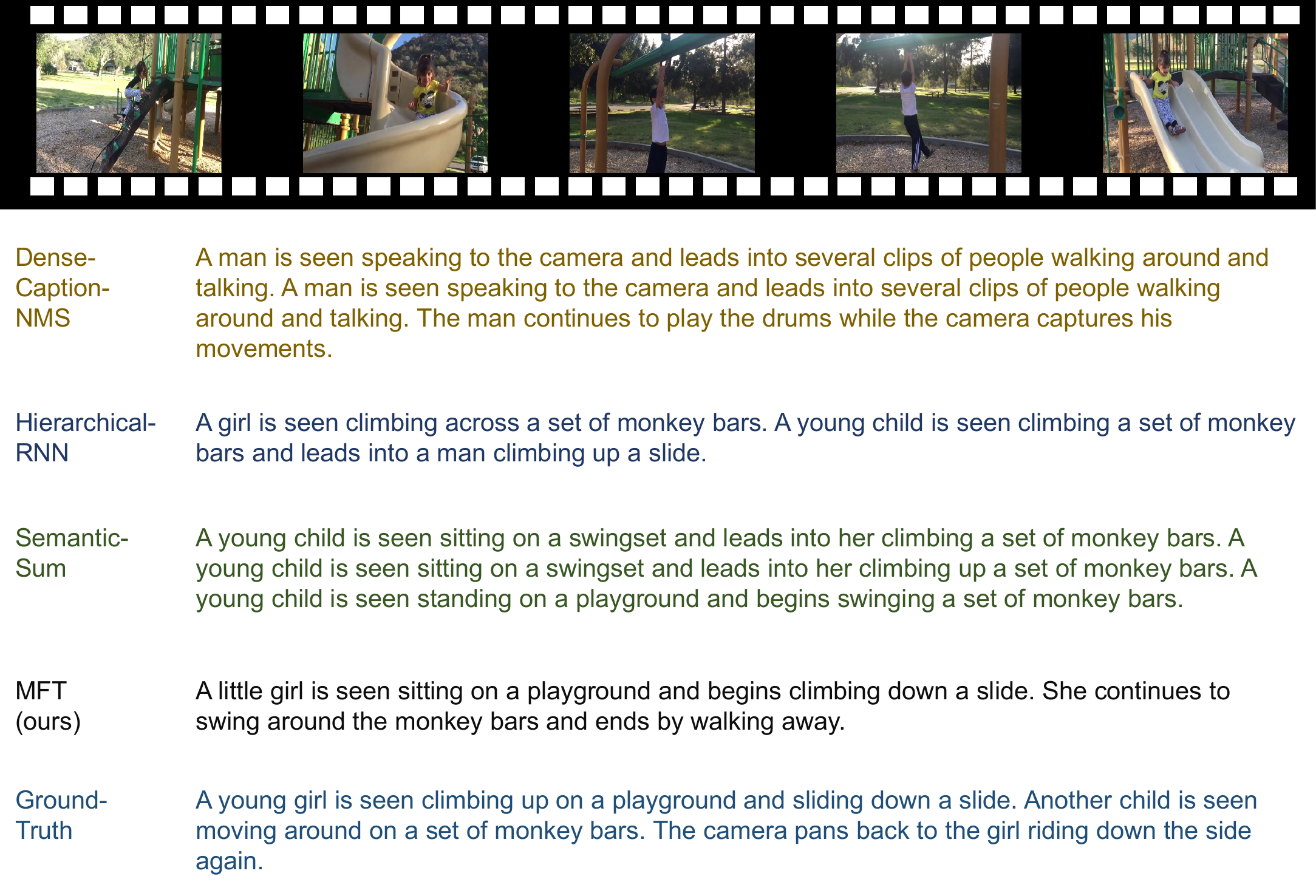}
\label{fig:sup3}
\end{minipage}

\begin{minipage}{\textwidth}
\centering
\includegraphics[width=\textwidth]{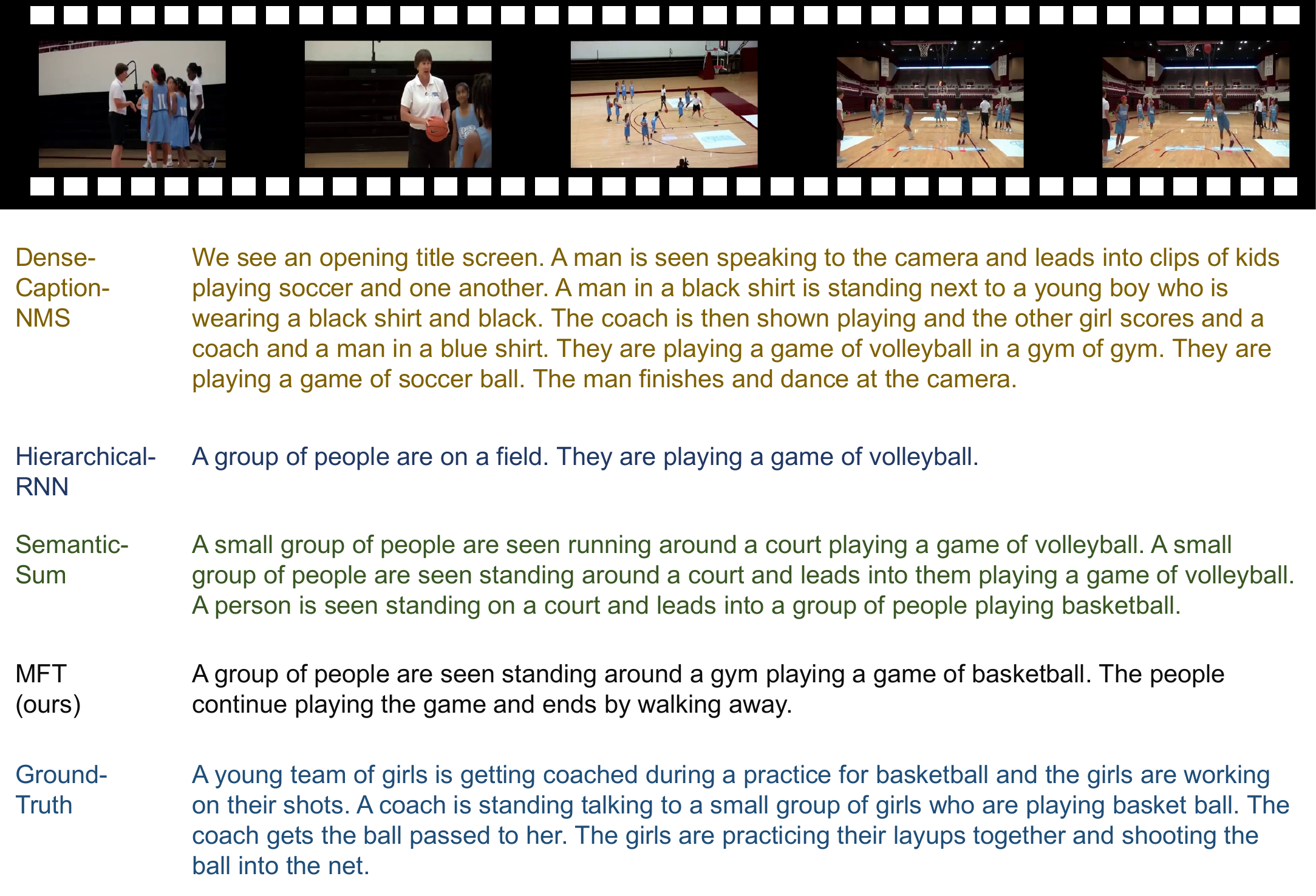}
\label{fig:sup4}
\end{minipage}
\caption{The third and the fourth samples. }
\end{figure}

\clearpage

\begin{figure}[h]
\centering
\includegraphics[width=\textwidth]{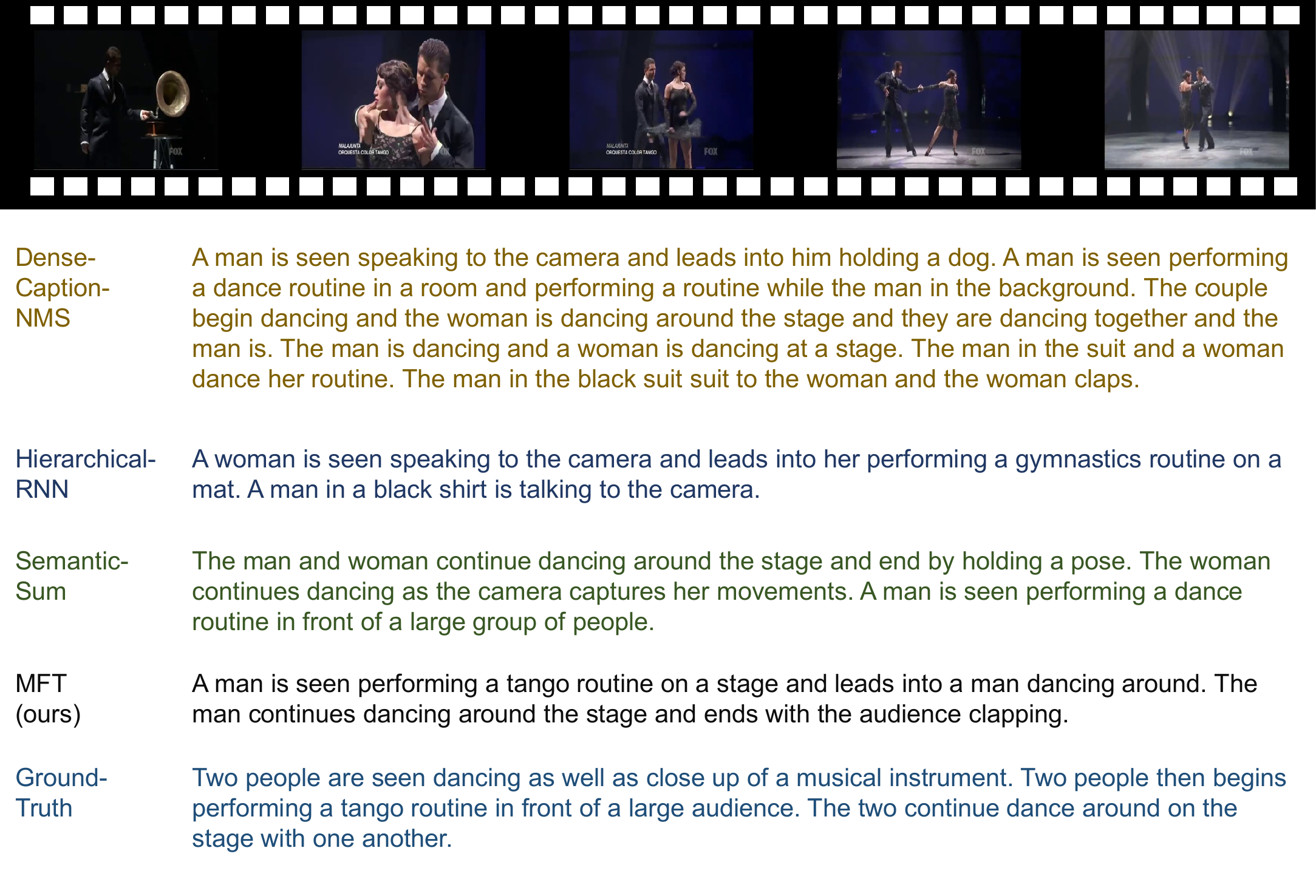}
\caption{The fifth sample.}
\label{fig:sup5}
\end{figure}

\end{document}